\PassOptionsToPackage{unicode}{hyperref}
\PassOptionsToPackage{hyphens}{url}
\PassOptionsToPackage{dvipsnames,svgnames,x11names}{xcolor}
\documentclass[
]{article}
\usepackage{amsmath,amssymb}
\usepackage{lmodern}
\usepackage{iftex}
\ifPDFTeX
  \usepackage[T1]{fontenc}
  \usepackage[utf8]{inputenc}
  \usepackage{textcomp} 
\else 
  \usepackage{unicode-math}
  \defaultfontfeatures{Scale=MatchLowercase}
  \defaultfontfeatures[\rmfamily]{Ligatures=TeX,Scale=1}
\fi
\IfFileExists{upquote.sty}{\usepackage{upquote}}{}
\IfFileExists{microtype.sty}{
  \usepackage[]{microtype}
  \UseMicrotypeSet[protrusion]{basicmath} 
}{}
\makeatletter
\@ifundefined{KOMAClassName}{
  \IfFileExists{parskip.sty}{%
    \usepackage{parskip}
  }{
    \setlength{\parindent}{0pt}
    \setlength{\parskip}{6pt plus 2pt minus 1pt}}
}{
  \KOMAoptions{parskip=half}}
\makeatother
\usepackage{xcolor}
\NewDocumentCommand\citeproctext{}{}
\NewDocumentCommand\citeproc{mm}{%
  \begingroup\def\citeproctext{#2}\cite{#1}\endgroup}
\makeatletter
 \let\@cite@ofmt\@firstofone
 \def\@biblabel#1{}
 \def\@cite#1#2{{#1\if@tempswa , #2\fi}}
\makeatother
\newlength{\cslhangindent}
\newlength{\csllabelwidth}
\newenvironment{CSLReferences}[2] 
 {\begin{list}{}{%
  \setlength{\itemindent}{0pt}
  \setlength{\leftmargin}{0pt}
  \setlength{\parsep}{0pt}
  \ifodd #1
   \setlength{\leftmargin}{\cslhangindent}
   \setlength{\itemindent}{-1\cslhangindent}
  \fi
  \setlength{\itemsep}{#2\baselineskip}}}
 {\end{list}}
\usepackage{calc}

\ifLuaTeX
\usepackage[bidi=basic]{babel}
\else
\usepackage[bidi=default]{babel}
\fi
\babelprovide[main,import]{american}

\def\languageshorthands#1{}
\ifLuaTeX
  \usepackage{selnolig}  
\fi
\IfFileExists{bookmark.sty}{\usepackage{bookmark}}{\usepackage{hyperref}}
\IfFileExists{xurl.sty}{\usepackage{xurl}}{} 
\hypersetup{
  pdftitle={Kleinkram: Open Robotic Data Management},
  pdfauthor={Cyrill Püntener, Johann Schwabe, Dominique Garmier, Jonas
Frey, Marco Hutter},
  pdflang={en-US},
  colorlinks=true,
  linkcolor={Maroon},
  filecolor={Maroon},
  citecolor={Blue},
  urlcolor={Blue},
  pdfcreator={LaTeX via pandoc}}

\title{Kleinkram: Open Robotic Data Management}

\definecolor{c53baa1}{RGB}{83,186,161}
\definecolor{c202826}{RGB}{32,40,38}

\usepackage[affil-it]{authblk}
\usepackage{orcidlink}
\author[1%
  *%
  \ensuremath\mathparagraph]{Cyrill Püntener%
    \,\orcidlink{0009-0000-5231-9310}\,%
    }
\author[1%
  *%
  \ensuremath\mathparagraph]{Johann Schwabe%
    \,\orcidlink{0009-0000-4490-0658}\,%
    }
\author[1%
  \ensuremath\mathparagraph]{Dominique Garmier%
    }
\author[1,2%
  \ensuremath\mathparagraph]{Jonas Frey%
    \,\orcidlink{0000-0002-7401-2173}\,%
    }
\author[1%
  ]{Marco Hutter%
    \,\orcidlink{0000-0002-4285-4990}\,%
    }

\affil[1]{Robotic Systems Lab, ETH Zurich, Switzerland%
  }
\affil[2]{Max Planck Institute for Intelligent Systems, Tübingen,
Germany%
  }
\affil[$\mathparagraph$]{Corresponding author: %
}
\affil[*]{These authors contributed equally.}
\date{16 June 2025}

\begin{document}
\maketitle

\section{Summary}\label{summary}

Data is key to advancing robotic perception, navigation, locomotion, and
reasoning. However, managing diverse robotic datasets presents unique
challenges, including scalability, quality assurance, and seamless
integration into diverse workflows. To address this gap, we introduce
Kleinkram, a free and open-source data management system tailored for
robotics research. Kleinkram enables efficient storage, indexing, and
sharing of datasets, from small experiments to large-scale collections.
It supports essential workflows like validation, curation, and
benchmarking via customizable compute actions. Designed with a
user-friendly web interface, CLI integration, and scalable deployment
options, Kleinkram empowers researchers to streamline data management
and accelerate robotics innovation.

\section{Statement of need}\label{statement-of-need}

To render robotic data useful for research, it is essential to store,
organize, and index the data in a way that makes it easily searchable
and shareable. While large corporations have developed internal tools or
rely on closed-source third-party providers, no openly available,
ready-to-use, and easy-to-deploy solution exists for the robotics
research community. Additionally, features such as data verification and
the ability to perform tailored compute jobs on newly generated datasets
are highly desirable, as they facilitate benchmarking, reproducibility
and algorithmic development.

To address these challenges, we introduce \textbf{Kleinkram}, an
on-premise cloud solution designed for scalable and efficient data
management. Unlike traditional cloud storage, Kleinkram natively
integrates compute capabilities, automates data transfer, and eliminates
the tedious manual effort typically associated with data management
workflows. By categorizing and structuring data around common robotics
use cases, Kleinkram facilitates the creation of large, diverse datasets
that can be easily shared and reused across multiple projects. Its
intuitive web interface ensures accessibility, while a command-line
interface (CLI) enables seamless integration into automated pipelines
and headless systems. Kleinkram supports widely adopted standards,
building on ROS1 and ROS2 message definitions, and offers native
compatibility with ROSbag and MCAP data formats.

\section{Data Structure}\label{data-structure}

Kleinkram is designed around the typical data generation process in
mobile robotics, assuming data is collected and stored primarily in
ROS1/ROS2-compatible ROSbag or MCAP file formats. Once data recording
for an experiment is complete, these files can be efficiently uploaded
and ingested into the Kleinkram system for centralized storage,
indexing, and subsequent post-processing. It is important to note that
the current version of Kleinkram focuses on post-recording and data
management. It does not support real-time data streaming or processing
on the fly.

To provide structure and facilitate organization and retrieval,
Kleinkram requires data to be organized according to a strict
three-layer hierarchy: Projects, Missions, and Files. Each layer
maintains a one-to-many relationship with the layer below it. While
users have flexibility in how they map their specific activities to this
structure, the intended model is that a Project represents a distinct
research project, which requires data storage. A Mission corresponds to
a single, self-contained experiment or data collection run conducted
within that project, and it contains all the individual data Files (like
ROSbag or MCAP files) recorded during that specific deployment. This
structured approach aids in navigating, managing, and understanding
large volumes of experimental data.

\section{System Architecture}\label{system-architecture}

Kleinkram's system architecture is modular, comprising several
interacting microservices.

\begin{itemize}
\item
  \textbf{Python Client Library and CLI} Provides programmatic access to
  Kleinkram's functionalities, enabling efficient data transfer
  operations (upload, download) directly from Python scripts or the
  command line. This allows seamless integration into robotic workflows
  and automated data pipelines running on robots or workstations,
  removing the need for manual browser interaction for data transfer.

  The CLI is built using the typer library, sharing a core Python
  codebase with the client library.
\item
  \textbf{Web interface} Serves as the primary graphical user interface
  for users to interact with Kleinkram. It allows for the browsing,
  managing and structuring files and their metadata.

  It is implemented as a single-page application using the Vue3
  framework and the Quasar component library.
\item
  \textbf{Backend API} Acts as the central communication layer between
  the client applications (web UI and Python client/CLI) and the data
  storage and processing components. It handles authentication, data
  indexing, metadata management, and schedules background tasks.

  The backend is implemented using the NestJS framework and utilises a
  PostgreSQL database for storing all metadata related to projects,
  missions, files, users, and actions.
\item
  \textbf{Data Store} The raw robotic data files (ROSbags, MCAPs) are
  stored on an S3-compatible object storage backend. This provides
  scalability and flexibility. Users can easily deploy and manage their
  own storage using self-hosted solutions like MinIO, or utilise public
  cloud S3 services. Kleinkram interacts with the data store via the S3
  API.
\item
  \textbf{Action Runner} This component enables the execution of
  customisable data processing and analysis tasks directly on the data
  stored within Kleinkram. Users can define ``Actions'' (e.g., validate
  data integrity, extract sensor metadata, generate preview
  visualisations, convert formats, run benchmarking scripts).

  These actions are packaged as Docker containers. The action runner
  orchestrates the execution of these containers, providing them access
  to the necessary data from the data store using the client library or
  CLI.
\item
  \textbf{Observability} (Optional) Monitoring and logging system
  performance, resource usage, and task execution status are crucial for
  managing a scalable data system. Integration with observability tools,
  such as the Grafana Stack (Loki for logs, Prometheus for metrics,
  Tempo for traces, Grafana for dashboards) can provide insights into
  the system's health and the progress of the data processing task.
\end{itemize}

\section{Usecase}\label{usecase}

Kleinkram has been used internally at the Robotic Systems Lab at ETH
Zurich over the past year. During this time, it has stored over 20 TB of
data collected from various robotic systems, effectively replacing the
lab's previous reliance on Google Drive for data storage. The largest
project supported by Kleinkram was the \textbf{GrandTour dataset}
(\citeproc{ref-frey25boxi}{Frey et al., 2025}), in which our legged
robot \textbf{ANYmal} (\citeproc{ref-hutter16anymal}{Hutter et al.,
2016}), equipped with \textbf{Boxi} (\citeproc{ref-frey25boxi}{Frey et
al., 2025}), a multi-sensor payload, was deployed across various
locations in Switzerland.

Following each data collection mission, raw data --- recorded in the
form of ROSbags and MCAP files --- was uploaded directly to Kleinkram
via its command-line interface (CLI). The intuitive Docker-based action
integration allowed us to easily define and execute data verification
tests. These include, for example, checks to ensure that all sensor
streams were recorded at the expected frequencies and correct time
synchronization was established during the distributed recordings, as
well as common sense checking for validity of data (e.g.~images are not
black or the IMUs measure the gravity vector).

Beyond data verification, Kleinkram enabled us to run full SLAM
pipelines retrospectively, automatically producing standard Absolute and
Relative Trajectory Error (ATE/RTE) metrics. This compute integration
was critical for development, benchmarking, and evaluation.

Equally important was Kleinkram's user-friendly CLI, which provided
quick access to summary statistics such as dataset counts, durations,
and other key metrics --- many of which were directly used in associated
publications. Given that for our use case, data has to be mainly
accessed within the ETH network infrastructure; datasets can be pulled
on demand and deleted afterwards, fully utilizing the fast on-premise
interconnect infrastructure without relying on external servers.

Throughout the project, Kleinkram also enforced metadata submission
during upload. Users were required to include a YAML file describing the
mission, which captured essential information such as the robot
operator, specific hardware configuration, location, and links to
related resources (e.g., associated Google Drive folders for images).
This structured metadata was essential for organizing and retrieving
data at scale.

\section{Acknowledgements}\label{acknowledgements}

This work was primarily supported by the Open Research Data Grant at ETH
Zurich. Jonas Frey is supported by the Max Planck ETH Center for
Learning Systems.

This work was supported and partially funded by Leica Geosystems, which
is part of Hexagon. In addition, this work was supported by the National
Centre of Competence in Research Robotics (NCCR Robotics), the ETH
RobotX research grant funded through the ETH Zurich Foundation, the
European Union's Horizon 2020 research and innovation program under
grant agreement No 101016970, No 101070405, and No 101070596, and an ETH
Zurich Research Grant No.~21-1 ETH-27.

We extend our sincere appreciation to Noel Kampus for the initial design
of the web interface, to Lars Leuthold and Marvin Lichtsteiner for their
valuable contributions, and to William Talbot and Turcan Tuna for their
efforts in the internal testing of Kleinkram.

\section*{References}\label{references}
\addcontentsline{toc}{section}{References}

\phantomsection\label{refs}

\end{document}